\documentclass[journal,twoside,web]{ieeecolor}
\usepackage{generic}
\usepackage{cite}
\usepackage{amsmath,amssymb,amsfonts}
\usepackage{algorithmic}
\usepackage{graphicx}
\usepackage{algorithm,algorithmic}
\usepackage{hyperref}
\hypersetup{hidelinks}
\usepackage{textcomp}
\usepackage{tabularx}
\usepackage{booktabs}
\usepackage{multirow}
\usepackage[normalem]{ulem}
\usepackage{soul}
\usepackage{placeins}
\usepackage{float}

\def\BibTeX{{\rm B\kern-.05em{\sc i\kern-.025em b}\kern-.08em
    T\kern-.1667em\lower.7ex\hbox{E}\kern-.125emX}}
\begin{document}
\title{Federated Foundation Model for GI Endoscopy Images}


\author{
Alina Devkota, Annahita Amireskandari, Joel Palko, Shyam Thakkar, Donald Adjeroh,\\
Xiajun Jiang, Binod Bhattarai, and Prashnna K. Gyawali
\thanks{Alina Devkota, Donald Adjeroh, and Prashnna K. Gyawali are with the Lane Department of Computer Science and Electrical Engineering, West Virginia University, Morgantown, WV, USA (e-mail: ad00139@mix.wvu.edu, donald.adjeroh@mail.wvu.edu, prashnna.gyawali@mail.wvu.edu)
(Corresponding author: Prashnna K. Gyawali)
}
\thanks{Annahita Amireskandari, Joel Palko, and Shyam Thakkar are with the School of Medicine, West Virginia University, Morgantown, WV, USA (e-mail: annahita.amireskandari@hsc.wvu.edu, joel.palko@hsc.wvu.edu, shyam.thakkar@hsc.wvu.edu)
}%
\thanks{Xiajun Jiang is with the Department of Computer Science, University of Memphis, Memphis, TN, USA (e-mail: Xiajun.Jiang@memphis.edu)
}%
\thanks{Binod Bhattarai is with the University of Aberdeen, UK (e-mail: binod.bhattarai@abdn.ac.uk)
}
}

\maketitle
\begin{abstract}
Gastrointestinal (GI) endoscopy is essential in identifying GI tract abnormalities in order to detect diseases in their early stages and improve patient outcomes. Although deep learning has shown success in supporting GI diagnostics and decision-making, these models require curated datasets with labels that are expensive to acquire. Foundation models offer a promising solution by learning general-purpose representations, which can be finetuned for specific tasks, overcoming data scarcity. Developing foundation models for medical imaging holds significant potential, but the sensitive and protected nature of medical data presents unique challenges. Foundation model training typically requires extensive datasets, and while hospitals generate large volumes of data, privacy restrictions prevent direct data sharing, making foundation model training infeasible in most scenarios. In this work, we propose a FL framework for training foundation models for gastroendoscopy imaging, enabling data to remain within local hospital environments while contributing to a shared model. We explore several established FL algorithms, assessing their suitability for training foundation models without relying on task-specific labels, conducting experiments in both homogeneous and heterogeneous settings. We evaluate the trained foundation model on three critical downstream tasks—classification, detection, and segmentation—and demonstrate that it achieves improved performance across all tasks, highlighting the effectiveness of our approach in a federated, privacy-preserving setting.
\end{abstract}


\vspace{-3pt}
\section{Introduction}
\vspace{-3pt}
\label{sec:introduction}

Gastrointestinal (GI) cancers represent a major global health burden, accounting for a substantial proportion of cancer cases and deaths worldwide—a trend projected to increase in the coming decades \cite{jha2023gastrovision}. Early detection through procedures such as endoscopy is critical for identifying GI tract abnormalities at a treatable stage. Enhancing endoscopic performance is therefore essential for reducing GI disease-related morbidity and mortality. GI vision tasks—such as polyp detection, inflammation classification, or early cancer diagnosis—demand robust, generalizable AI models trained on large, diverse datasets. With the growing success of artificial intelligence (AI) in supporting GI diagnostics and decision-making \cite{pannala2020artificial}, there is a pressing need to develop scalable models tailored for GI endoscopy that can generalize across clinical settings.

Deep learning has revolutionized medical image analysis, including GI endoscopy \cite{zhang2023effect, pannala2020artificial}, by enabling automated detection, classification, and segmentation tasks with remarkable accuracy. The field has witnessed a rapid evolution—from early convolutional neural networks (CNNs) \cite{shin2016deep} to deeper architectures like ResNets \cite{he2016deep}, and attention-based models such as Vision Transformers (ViTs) \cite{dosovitskiy2020image}. 
These increasingly large and expressive models, when paired with large labeled datasets, have achieved state-of-the-art performance across diverse applications. In the GI domain, task-specific deep learning models have been developed for detecting polyps \cite{mamonov2014automated}, segmenting lesions \cite{ali2021deep}, and classifying inflammation severity \cite{fan2023novel}. 
However, these models are typically trained in a fully supervised manner, requiring high-quality annotated datasets tailored to each specific task, which can be prohibitively expensive and time-consuming to acquire in clinical settings.
This is particularly true for GI endoscopy, where data collection involves costly procedures, expert endoscopists' time, and labor-intensive frame-by-frame labeling of long endoscopic videos.
Moreover, their task-specific nature often limits their generalizability to new or slightly different domains.

To address these limitations, there has been recent traction toward building Foundation Models (FMs)—large, pretrained models trained on diverse data using task-agnostic objectives, which can then be adapted to a variety of downstream tasks with minimal supervision \cite{wang2023foundation, zhou2023foundation}. 
FMs operate by learning rich general-purpose representations that capture semantic, structural, or temporal patterns across datasets \cite{zhou2023foundation} using objectives such as masked autoencoding \cite{he2022masked} and contrastive learning \cite{chen2020simple}, without relying on dense annotations.
In GI endoscopy, FMs hold significant promise for overcoming data scarcity, reducing annotation burdens, and unifying multiple GI vision tasks such as polyp detection, mucosal characterization, and lesion segmentation under a single pretrained backbone. This paradigm shift opens the door to building more flexible and generalizable models for GI applications.

While FMs offer a promising paradigm for building generalizable and annotation-efficient solutions, their success hinges on access to large and diverse datasets \cite{awais2025foundation}. In the medical domain, however, such data is often siloed across institutions due to privacy concerns, differing data governance policies, and the sensitive nature of patient information \cite{adnan2022federated}, making centralized data aggregation particularly challenging. 
To address these constraints, Federated Learning (FL) has emerged as a compelling solution, enabling models to be trained collaboratively across distributed clients—such as hospitals or clinics—without sharing raw data \cite{mcmahan2017communication,  thrasher2023multimodal}. Instead, only model updates are exchanged, preserving patient confidentiality while leveraging a broader pool of data. However, most existing FL applications have been limited to task-specific, supervised training, such as local classification or segmentation tasks. There has been limited exploration of federated foundation learning, where task-agnostic, large-scale pretraining is conducted across data silos to build general-purpose models. 
Unlocking this potential could lead to more robust, scalable, and equitable AI systems for GI vision tasks, trained in a privacy-preserving and collaborative manner.

To this end, we propose Federated Foundation Models (FFMs) to learn general-purpose representations for gastrovision tasks.
Without requiring labels, we train a foundation model in 
a federated
fashion using a masked autoencoding objective, enabling the collaborative construction of a strong foundational representation in a setting where unlabeled gastroenterological imaging datasets reside in silos.
To reflect real-world variability in imaging protocols, devices, and patient populations, we rigorously evaluate our approach under both homogeneous (uniform data distributions) and heterogeneous (non-i.i.d. data across clients) scenarios. In the heterogeneous setting, we simulate real-world scenarios where clients have imbalanced and non-overlapping distributions of GI regions and case types, with some entirely missing at certain sites, highlighting the importance of building adaptable and robust models in federated environments.

After training the FFM, we fine-tune it on three downstream gastrovision tasks: classification, detection, and segmentation.
We benchmark its performance against two non-federated baselines: a lower-bound setup (local training without collaboration) and an upper-bound setup (centralized training by aggregating all data), an ideal but impractical scenario due to privacy constraints.
While the upper-bound represents best-case performance, our experiments show that the FFM consistently approaches it, significantly outperforming the lower-bound across all tasks.
These results highlight the effectiveness of our federated foundation modeling in delivering strong generalization and transferability without compromising data privacy—demonstrating its practical utility for real-world gastro vision applications. 
Overall our contributions are:

\begin{itemize}
    \item We train FFMs for GI endoscopy datasets and evaluate them on classification, detection, and segmentation tasks. Our results demonstrate the effectiveness of FFMs in generalization without compromising data privacy.
    \item We evaluate our methods on both homogeneous and heterogeneous scenarios and demonstrate the applicability of federated pretraining in real-life conditions.
    \item Our convergence analysis shows that FFMs can achieve stable and near-optimal learning dynamics.
\end{itemize}

\vspace{-3pt}
\section{Related Work}
\vspace{-3pt}
\subsection{Deep Learning and Foundation Models in GI Endoscopy}
Deep learning has significantly advanced GI endoscopy by enhancing diagnostic accuracy and supporting clinical decision making \cite{pannala2020artificial, sharma2023deep}. 
It has demonstrated utility in improving tasks like disease classification \cite{pannala2020artificial}, lesion detection \cite{mamonov2014automated, yamada2019development}, and segmentation \cite{rahman2024attention, ali2021deep}. As the standard deep learning methods matured, there was a shift towards exploring the benefits of transfer learning and pretraining to improve performance, particularly in data-scarce clinical settings. For example, \cite{sharma2023deep} emphasized the diagnostic value of using pretrained models in GI imaging, and \cite{de2020deep} pretrained a hybrid ResNet-UNet on endoscopic images. Recently, \cite{boers2024foundation} highlighted the importance of in-domain pretraining for GI datasets.

Despite the success of standard deep learning models in GI endoscopy, they are highly task-specific and often tailored to narrow imaging modalities, limiting their generalizability across diverse datasets and downstream tasks. To address these challenges, FMs have emerged as a promising solution.
For example, \cite{wang2023foundation} trained an FM from unlabeled endoscopic video data, and
EndoDINO \cite{dermyer2025endodino} and EndoViT \cite{batic2024endovit}, pretrained on endoscopic images, have been applied to a range of downstream tasks like lesion detection, anatomic landmark classification, tissue and polyp segmentation, surgical phase recognition, etc. 
However, the challenges lie in the development of FMs due to data heterogeneity and data privacy, regulatory, and ethical concerns.
In this study, we aim to address these challenges by developing privacy-preserving foundation models for GI endoscopy to realize the full potential of FMs in healthcare.

\subsection{Federated Representation Learning}
Recent works in self-supervised FL has focused on improving representation learning without requiring labeled data.
FedEMA \cite{zhuang2022divergence} and Hetero-SSFL \cite{makhija2022federated} demonstrated the capability of self-supervised FL even under heterogeneous client distributions. \cite{yan2023label} presents another self-supervised FL framework for medical image analysis, that improved model robustness against various degrees of data heterogeneity. FMs are often pre-trained in self-supervised settings, and these works demonstrate the capabilities of Federated self-supervised representation learning.
Additionally, there have been some attempts to combine FL and FMs to develop more robust, efficient, inclusive, and privacy-preserving AI models. FedFMS \cite{liu2024fedfms} is an example of FFM for medical image segmentation, which achieved performances of centrally trained models while maintaining privacy.
\cite{zhou2025federated} provides an in-depth evaluation of FL algorithms applied to medical image classification tasks. In GI endoscopy, 
Federated EndoViT \cite{kirchner2025federated} used FL to train foundation models for surgical data analysis. The existing works on FFMs for medical images are either task-specific \cite{liu2024fedfms, zhou2025federated} or explore a single federated configuration \cite{kirchner2025federated}. In our work, we develop FFMs tailored for GI endoscopy images and demonstrate effective fine-tuning across multiple downstream tasks. Additionally, we evaluate our methods on both homogeneous and heterogeneous client distributions to demonstrate their practical utility.

\vspace{-3pt}
\section{Methodology}
\vspace{-3pt}

\begin{figure}
	\centerline{\includegraphics[scale=.15]{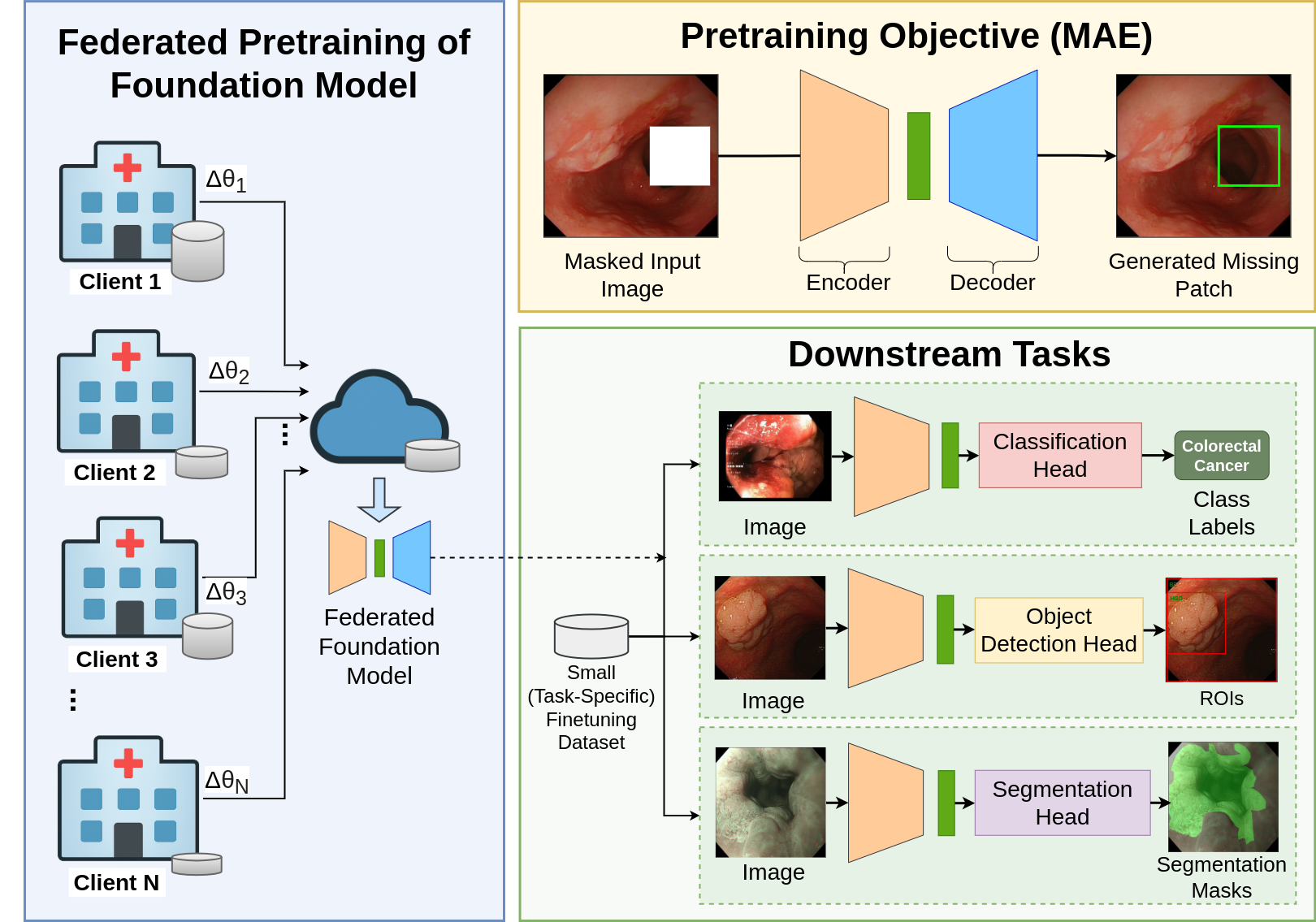}}	
	\caption{Workflow of Federated Foundation Model. The federated pretraining setup consists of multiple clients and a central server. Each client trains its local model using the local dataset and computes model updates and sends them to the server to form a global model ($\theta$) via aggregation. The pretrained FFM is fine-tuned on three downstream tasks: classification, object detection, and semantic segmentation.}
	\label{fig:methodology}
\end{figure}

In this section, we first provide background on foundation models and federated learning, and then describe our methodology for federated foundation models. Fig. ~\ref{fig:methodology} presents an overview of our approach, illustrating the federated pretraining of the foundation model and the downstream fine-tuning across three different tasks in GI endoscopy analysis.

 \vspace{-3pt}

\vspace{-3pt}
\subsection{Preliminary I: Foundation Model}
\vspace{-3pt}
Foundation models (FMs) are large-scale models trained on broad data distributions that can be adapted to a variety of downstream tasks with minimal task-specific modifications \cite{wang2023foundation}.
They are typically pretrained in a self-supervised or unsupervised manner on massive datasets to learn general-purpose representations.
Mathematically, an FM $\mathcal{M}$ with parameters $\theta$ is trained to minimize a loss function $\mathcal{L}$ defined over diverse pretraining objectives. 
Let $X \in \mathbb{R}^d$ denote the input data and $\widetilde{X}$ represent a target derived from $X$ based on the chosen objective—such as an augmented view in contrastive learning, a masked version in masked modeling, or the original input itself in reconstruction-based learning.
The training objective, $\mathcal{L}_{\text{pretrain}}$, is then formulated as:
\vspace{-3pt}
\begin{equation}
\label{eqn:fm}
    \mathcal{L}(\theta^*) = \arg\min_{\theta}  \mathbb{E}_{X} \left[ \mathcal{L}_{\text{pretrain}}(\mathcal{M}(X; \theta), \widetilde{X}) \right]
\end{equation} \vspace{-3pt}
We use the Masked Autoencoder (MAE) framework with L1 loss to pretrain our FM on a large corpus of unlabelled endoscopy images. The MAE approach enables the model to learn meaningful representations by reconstructing missing patches of the input images, without relying on explicit labels.

After completing pretraining, the learned representations $z=\mathcal{M}(X;\theta^*)$ can either be used directly for downstream tasks or the model $\mathcal{M}$ can be fine-tuned for specific objectives. During fine-tuning, the model's parameters $\theta^*$ are updated to minimize a task-specific loss function using a labeled dataset. Let $\{(X_i, y_i)\}_{i=1}^N$ denote the fine-tuning dataset, where $X_i$ is the input and $y_i$ is the corresponding ground-truth label. The fine-tuning objective is given by:
\begin{equation}
\label{eq:finetuning}
    \mathcal{L}_f (\hat{\theta}) = \sum_{i=1}^{N} \mathcal{L}_{\text{task}}\left( \mathcal{M}(X_i; \theta^*), \mathbf{y}_i \right)
\end{equation}
where $\mathcal{L}_{\text{task}}$ is loss function specific to the downstream task. 
It may be the cross-entropy loss for classification, Dice loss for segmentation, or focal loss for object detection.
We finetune our pretrained models on three distinct downstream tasks—classification, object detection, and segmentation. Performance on these tasks reflect the capacity and generalizability of the FMs to domain-specific applications.

\vspace{-3pt}
\subsection{Preliminary II: Federated Learning}
\vspace{-3pt}
Federated learning (FL) \cite{mcmahan2017communication} leverages decentralized data sources by collaboratively training models across multiple devices or nodes.
Instead of sharing raw
data, clients exchange model parameters to build a global model shared among all nodes.
In FL, each of the $K$ nodes downloads an initial model $\theta_0$ from the central server, trains the local model $\theta_k$ locally using its local data $X_k$ by optimizing the local objective $F_k$, and sends the model updates back to the server. 
At the end of each communication round $r$, the central server 
aggregates the locally trained models to form an updated global model $\theta_r$, which is then redistributed to all nodes.
This process repeats iteratively until convergence.
The overall goal of FL is to minimize a global objective function $f$, which is typically expressed as a weighted sum of the local objectives:
\vspace{-3pt}
\begin{equation}
    \label{eqn:fed}
    \min_{w} f(w) = \frac{1}{n} \sum_{k=1}^{K} n_k F_k(w)
\end{equation} \vspace{-3pt}
where $n_k$ is the number of data points at node $k$, and $n = \sum_{k=1}^{K} n_k$ is the total number of data points across all nodes.


\vspace{-3pt}
\subsection{Federated Foundation Model}
\vspace{-3pt}
Conventional FM training relies on large and centralized datasets, which are often impractical in healthcare, where data, such as GI endoscopy images, are distributed and privacy-sensitive.
To address these limitations, we propose
Federated Foundation Model (FFM) — a distributed pretraining framework for foundation models that enables collaborative learning across institutions without sharing raw data. Unlike traditional setups where the entire training corpus is aggregated on a single server, FFM distributes the data across 
$K = N + 1$ nodes, comprising 
$N$ client nodes and one central server.

In our setting, each client node holds a private and unlabeled dataset $D_1$, $D_2$, $D_3$, and $D_N$, while the server node also possesses a local dataset $D_s$. This contrasts with standard FL setups, where only client nodes contribute data. The inclusion of server-side data allows us to study a novel and meaningful benchmark: how collaborative training with clients improves over isolated training on server data alone, and how close this performance approaches that of a centrally trained FM.

Each client and the central server first train their models locally on their respective datasets and compute model updates \(\Delta \theta_1\), \(\Delta \theta_2\), \(\Delta \theta_3\), ..., \(\Delta \theta_N\) and \(\Delta \theta_s\).
The server then aggregates these model updates to form a global model $\mathcal{M}$ with parameters $\theta$, which is subsequently distributed to all nodes.
This distributed setup for pretraining FMs allows us to leverage data located across multiple, potentially privacy-sensitive sites. In our framework, the pretraining of FMs is conducted using FL, and fine-tuning for downstream tasks is performed on a single node, reflecting how individual hospitals might independently leverage FMs for their respective objective. Using the formulation defined earlier in Eqn. \eqref{eqn:fm} and Eqn. \eqref{eqn:fed}, the global objective for the proposed FFM becomes:
\vspace{-3pt}
\begin{equation}
\label{eq:ffm-pretrain1}
\theta^* = \arg\min_{\theta} \frac{1}{n} \sum_{k=1}^{K} n_k \, \mathbb{E}_{X_k \sim \mathcal{D}_k} \left[ \mathcal{L}_{\text{pretrain}}^k\left(\mathcal{M}(X_k; \theta_k), \widetilde{X_k} \right) \right]
\end{equation}
Eqn. \eqref{eq:ffm-pretrain1} defines the overall federated pretraining objective, where ${\mathcal{L}_{\text{pretrain}}^k}$ is the pretraining loss function at node $k$, $X_k$ is the local input data, and $\widetilde{X_k}$ is the corresponding self-supervised target (e.g., masked patches in MAE). In our case, we adopt the MAE approach, where ${\mathcal{L}_{\text{pretrain}}^k}$ is L1 reconstruction loss. The global model, as shown in Eqn. \eqref{eq:ffm-pretrain1}, is formed by aggregating the locally trained FM models.

\textbf{Federated Learning Algorithms:}
While Eqn. \eqref{eq:ffm-pretrain1} illustrates a standard aggregation method, i.e., FedAvg \cite{mcmahan2017communication}, our proposed FFM can incorporate a wide range of FL strategies. 
This adaptability allows us to explore and evaluate different aggregation mechanisms, which we demonstrate in the experimental section.
Popular alternatives such as FedAvgM \cite{hsu2019measuring}, FedAdam \cite{reddi2020adaptive}, and FedAdagrad \cite{reddi2020adaptive} can also be employed within our framework without modification.
Algorithm \ref{algo:fed} (Supplementary) demonstrates the working of our FFM for various FL algorithms.
For clarity, we refer to FFMs trained using FedAvg as FedFoundAvg. Similarly, models trained using FedAvgM, FedAdam, and FedAdagrad are referred to as FedFoundAvgM, FedFoundAdam, and FedFoundAdagrad, respectively.
We use FedFound to collectively refer to these FFMs.


\textbf{Finetuning FFM:} For downstream tasks, we fine-tune the pretrained FFM by extracting features from the MAE encoder and attaching task-specific heads. This step enables the model to adapt the general representations learned during federated pretraining to specific supervised objectives. For classification, we attach an MLP head to the encoder. For object detection and instance segmentation, we adopt the ViTDet architecture \cite{li2022exploring}, replacing its standard feature extractor with our MAE encoder. Fine-tuning is performed using the objective defined in Eqn.~\eqref{eq:finetuning}, where each task employs an appropriate loss function: cross-entropy for classification, and Mask-RCNN \cite{he2017mask} based multi-task loss (classification loss + bounding box regression loss + Hungarian matching cost) for detection, and mask loss (Dice + BCE loss) in addition to Mask-RCNN based multi-task loss for segmentation.

\begin{figure}[!t]
\centerline{\includegraphics[scale=.4]{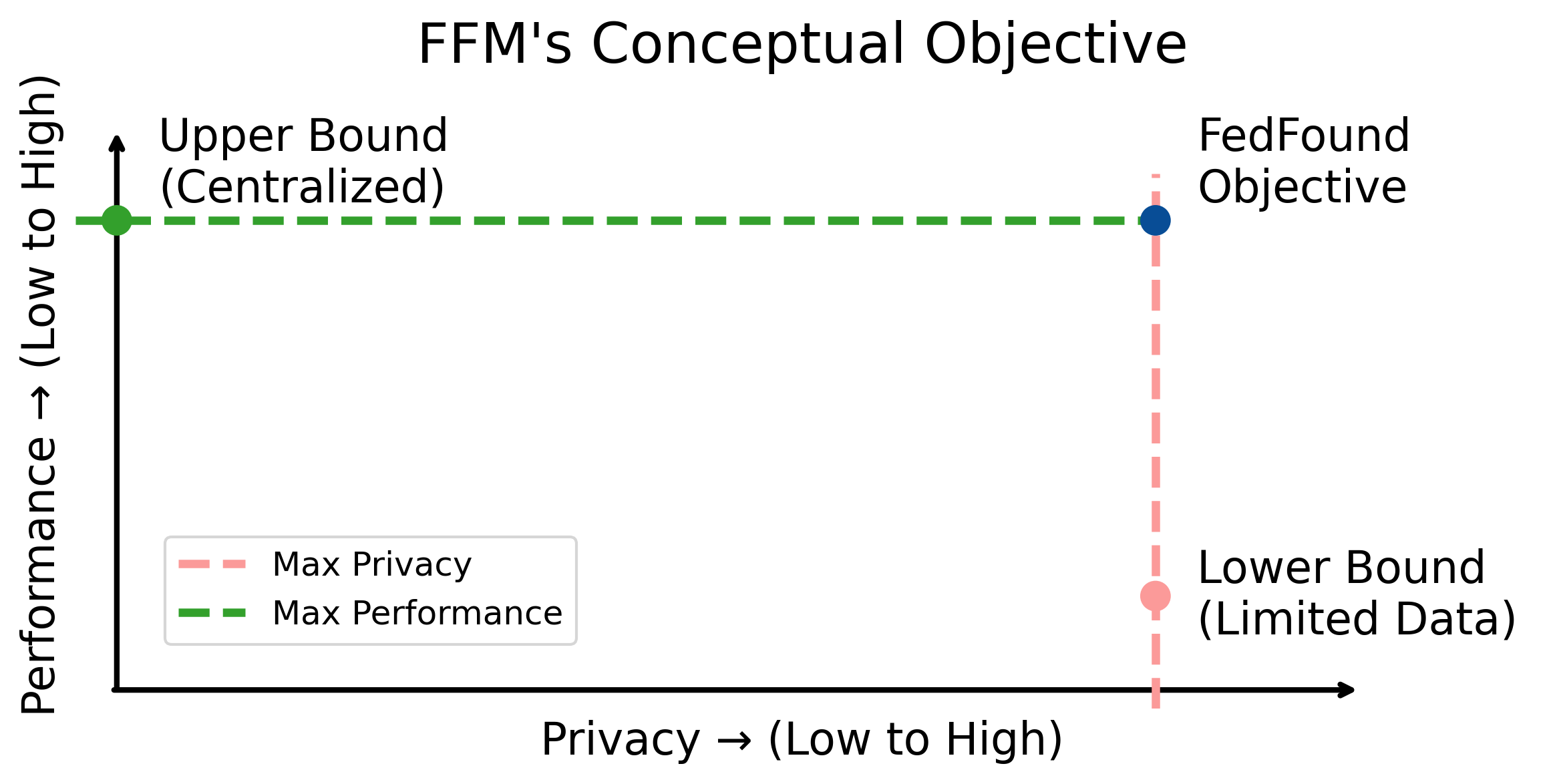}}	
	\caption{Conceptual trade-off between privacy and performance in foundation model training. Our federated training objective aims to achieve high performance with strong privacy, bridging the gap between centralized (upper bound) and single-node (lower bound) training.}\vspace{-3pt}
	\label{fig:objective}
\end{figure}

\textbf{FFM's Conceptual Objective:} As presented so far, FFM is designed to bridge the performance gap between two extremes:
(1) pretraining on a limited, non-representative dataset from a single institution (e.g., the server node), and
(2) pretraining using centralized access to all datasets, which poses significant privacy risks.
By aggregating knowledge from distributed nodes, FFM mitigates the privacy concerns of centralized training and simultaneously enhances model generalizability. To formalize this idea, we introduce the notion of \textit{upper bound} and \textit{lower bound} performance in the context of FFM training. The upper bound represents the ideal centralized scenario where all data is accessible for training, while the lower bound corresponds to training on a single institution's dataset, often limited and non-representative. 

Our objective is not only functional — improving FM training under data-distributional constraints — but also conceptual, as it redefines a meaningful benchmark for federated pretraining. This conceptual framework is illustrated in Fig. ~\ref{fig:objective}, where the trade-off between performance and privacy is visualized. The upper bound (centralized training) is expected to yield the highest performance (higher on the performance axis) but the lowest privacy (lower on the privacy axis). Conversely, the lower bound (single-institution training) ensures maximum privacy but typically results in suboptimal performance. FFM seeks to achieve a favorable balance between these extremes—retaining strong privacy guarantees while approaching the performance of centralized models.

This conceptual framework not only guides our evaluation but also provides a valuable benchmarking setup for future research in FFM training. By explicitly defining upper and lower bounds in terms of privacy and performance, it enables systematic comparison of new methods, architectures, and aggregation strategies within a common reference space. We believe this setup can serve as a standardized baseline to assess progress toward privacy-preserving, high-performing FMs in sensitive domains like healthcare.

\vspace{-3pt}
\section{Experiments}
\vspace{-3pt}
In this section, we present the experimental details, including a description of the datasets, model architecture, training setup, and evaluation protocol of this study.
\vspace{-3pt}
\subsection{Datasets}
\vspace{-3pt}
\label{sec:datasets}

\begin{figure}[!t]
	\centerline{\includegraphics[scale=.24]{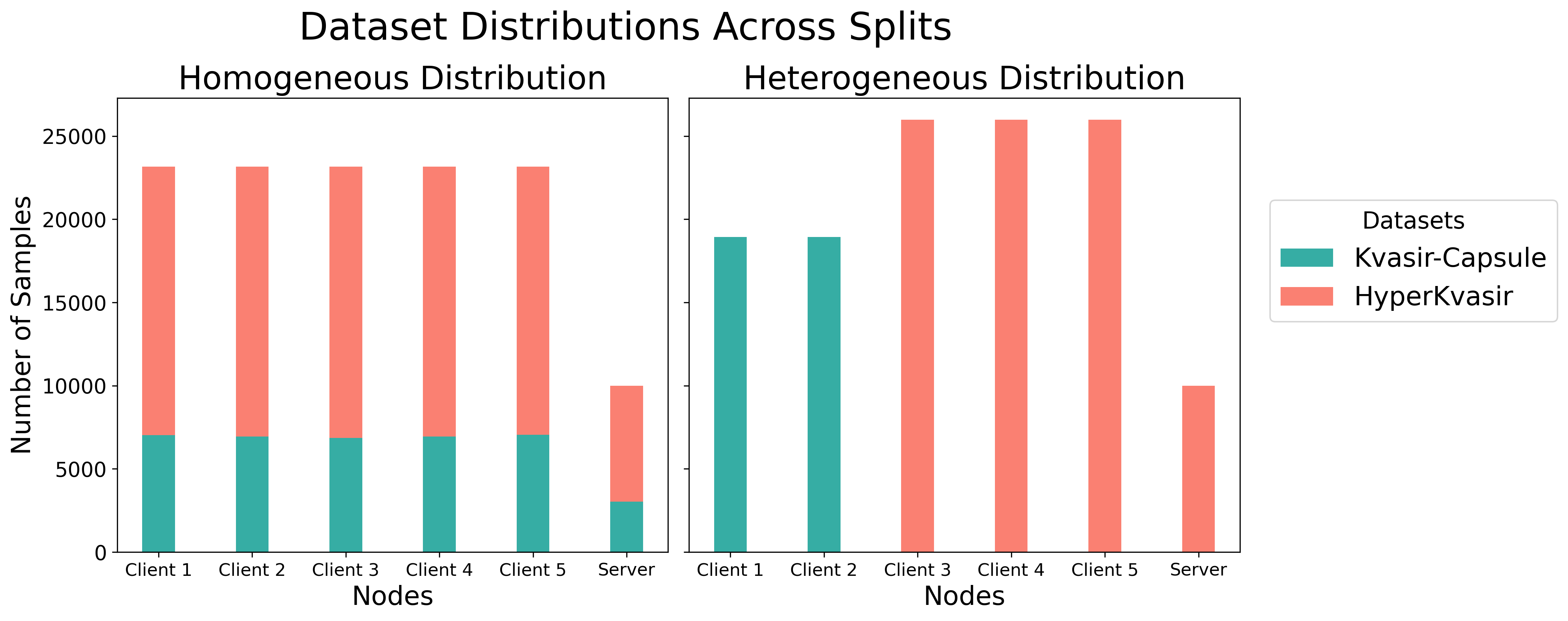}}
		
	\caption{Image distribution across homogeneous and heterogeneous splits for federated pretraining. The bar chart shows the number of images assigned to each node (Client 1 to Client 5 and Server) using different homogeneous and heterogeneous distribution strategies.}\vspace{-3pt}
	\label{fig:splits}
\end{figure}

We pretrain our foundation models using HyperKvasir \cite{borgli2020hyperkvasir} and Kvasir-Capsule \cite{smedsrud2021kvasir} datasets, both collected at Bærum Hospital, Norway. HypeKvasir dataset was collected from 2008 to 2016 using standard endoscopy equipment, whereas Kvasir-Capsule was collected from 2016 to 2018 using Endocapsule. These datasets also differ from one another in terms of image resolution, focus region within the GI tract, and the types of clinical findings they represent.
HyperKvasir consists of 110,079 endoscopy images, and Kvasir-Capsule consists of 47,238 images; from these, 87,970 and 37,876 images were randomly selected, respectively, for model pretraining.

Finetuning was conducted on Gastrovision \cite{jha2023gastrovision} and EDD2020 \cite{f8xg-wb80-20} datasets. The Gastrovision dataset consists of 8,000 images acquired from Bærum Hospital, Norway, and Karolinska University Hospital, Sweden, obtained using standard endoscopy equipment. It consists of 27 different classes, which include anatomical landmarks, pathological findings, polyp removal cases, and normal findings, and was used for the classification task. The EDD2020 dataset comprises 386 images spanning five disease/pathological classes, collected from different hospitals in France, Italy, and the UK. It is a multiclass dataset containing bounding boxes and segmentation masks, and was therefore used for downstream tasks of object detection and semantic segmentation. 



\textbf{Construction of homogeneous and heterogeneous data splits for pretraining:}
We split the pretraining images from the HyperKvasir and Kvasir-Capsule datasets across six nodes to create both homogeneous and heterogeneous splits. 
For the homogeneous split, we pooled 87,970 images from HyperKvasir and 37,876 images from Kvasir-Capsule, then randomly assigned 18,938 images to each of the five clients and 10,000 images to the server.
For the heterogeneous split, we randomly divided the Kvasir-Capsule images for two client nodes. 
To create the server dataset, we randomly sampled 10,000 images from HyperKvasir and distributed the remaining images equally among the remaining three clients.
In the homogeneous distribution, both datasets are evenly split across nodes, whereas in the heterogeneous distribution, the splits are uneven—some nodes receive significantly more data than others, and from different datasets.
While there may be alternative ways to create heterogeneous splits (e.g., different ratios of two datasets in each client), we believe this strategy is more realistic as different client nodes (e.g., hospitals) might have samples from different data. It also reflects an extreme non-iid scenario with domain shift due to differences in imaging procedure, focus region, and temporal structure.
This strategy allows us to simulate realistic FL scenarios in the absence of standardized benchmarks for constructing homogeneous and heterogeneous data splits, thereby providing a flexible and reproducible evaluation setup.
The distribution of images across clients and the server is shown in Fig. \ref{fig:splits}.

\vspace{-3pt}
\subsection{Model Architecture and Training}
\vspace{-3pt}
We use an MAE architecture for FFM, with 
a 12-layer encoder having 12 attention heads and an embedding dimension of 768, and an 8-layer decoder with 16 attention heads and an embedding dimension of 512.
Input images are resized to 256×256 with a patch size of 16, and a masking ratio of 0.6 is applied during pretraining.
The model was trained for 1,000 epochs — a value selected based on performance on downstream tasks — using a batch size of 256 on 4 NVIDIA A100 GPUs. Training the lower-bound model took approximately 6 hours, whereas federated pretraining and upper-bound training took 2 days and 15 hours due to sequential training imposed by resource constraints.
For downstream tasks, we use the pretrained encoder with an MLP head for classification, and replace the ViT backbone in ViTDet \cite{li2022exploring} for detection and segmentation. 
The fine-tuning tasks were resource-efficient. Classification on the Gastrovision dataset took $\sim$ 1 hour and 40 minutes (batch size 64) on two NVIDIA A30 GPUs. Object detection and segmentation required $\sim$ 3 and $\sim$ 4 hours, respectively, with a batch size of 8 on an NVIDIA A40 GPU. This computational efficiency of the downstream tasks demonstrates the practicality and scalability for real-world applications where frequent fine-tuning needs to be performed.

\vspace{-3pt}
\subsection{Evaluation}
\vspace{-3pt}
We evaluated the FFMs on multiple downstream tasks to assess their generalizability, using standard protocols for consistency and comparability with prior works.
For classification, we report the accuracy of the fine-tuned model. For object detection, we use mean Average Precision (mAP) at different Intersection over Union (IoU) thresholds—specifically, mAP@50 (IoU = 0.5) and mAP@75 (IoU = 0.75). For segmentation tasks, we report mAP over IoU thresholds, along with the Dice Score (DSC) and Jaccard Coefficient (JC) to measure the overlap between predicted and ground truth masks.
As described earlier (in section \ref{sec:datasets}), classification is evaluated on the GastroVision dataset, while object detection and segmentation are evaluated on the EDD2020 dataset.

\vspace{-3pt}
\section{Results}
\vspace{-3pt}

\begin{figure}[!t]
	\centerline{\includegraphics[scale=.25]{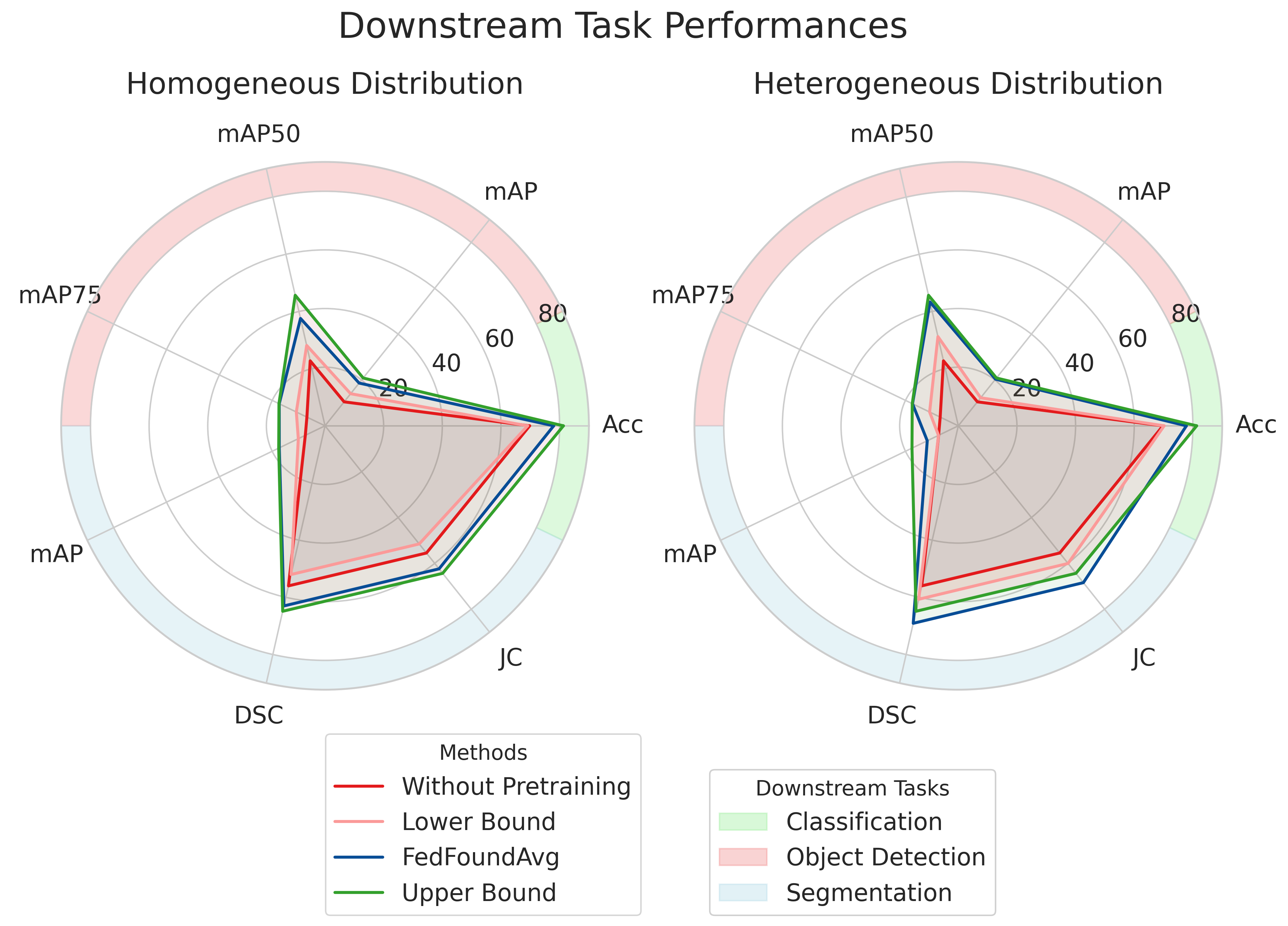}}	
	\caption{Comparison of foundation model performance across downstream tasks (classification, object detection, and segmentation) under homogeneous (Left) and heterogeneous (Right) client distributions.}
	\label{fig:results_overall_split1}
\end{figure} 

In this section, we present the detailed performance of the FFMs across various downstream tasks, highlighting its effectiveness in classification, object detection, and semantic segmentation benchmarks.
We report primarily results for four categories of federated foundation models: FedFoundAvg, FedFoundAvgM, FedFoundAdam, and FedFoundAdagrad, corresponding to different aggregation strategies used during training. To contextualize performance, we also compare these models against two reference points: a lower-bound centralized model trained on a single client and an upper-bound model trained with fully centralized data.

\vspace{-3pt}
\subsection{Performance on Downstream Tasks for Homogeneous Distribution}
\vspace{-3pt}

\begin{figure}[!t]
	\centerline{\includegraphics[scale=.23]{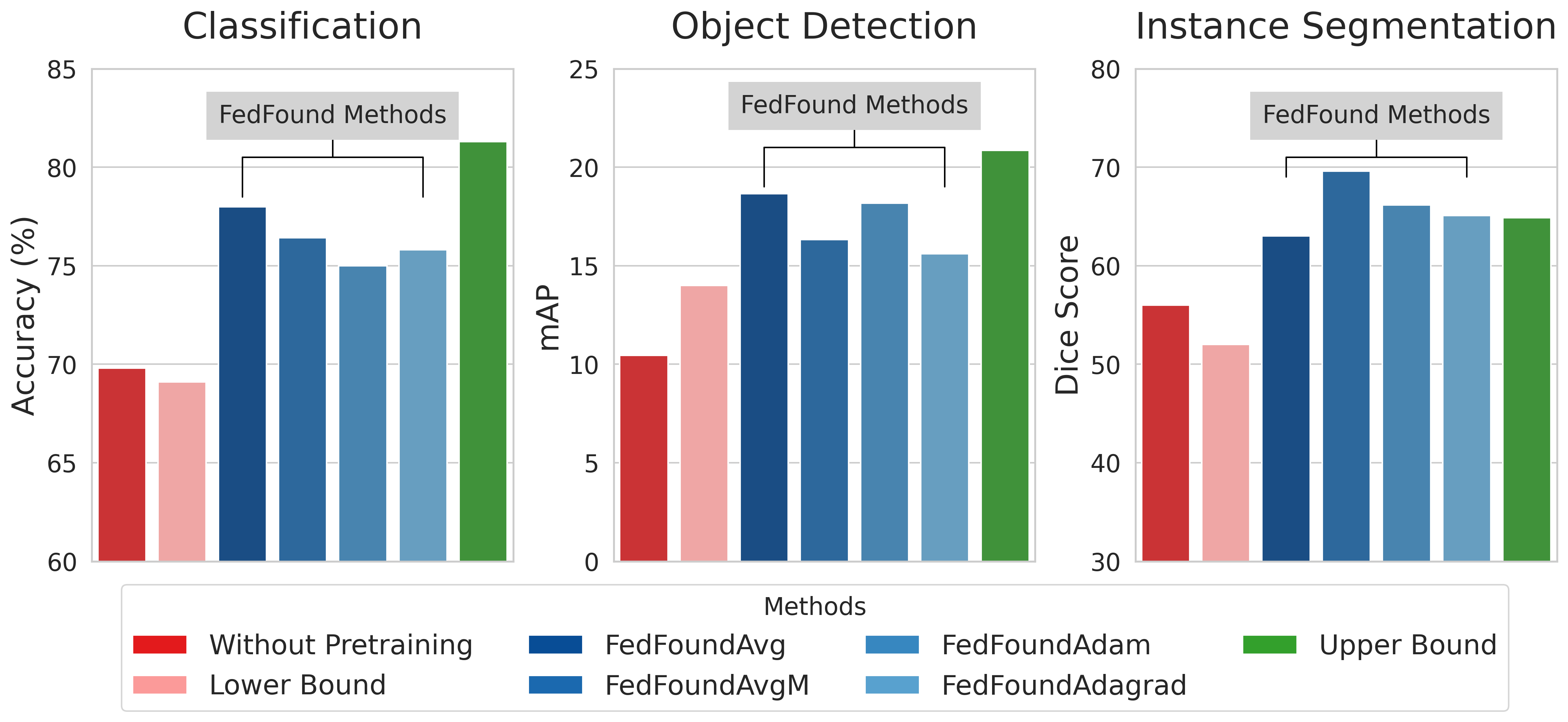}}
	\caption{Performance comparison of various methods on the homogeneous split. FedFound Models are evaluated against baseline, lower bound, and upper bound models across different downstream tasks.}
	\label{fig:results_homogeneous}
\end{figure}

We present the main results in Fig. \ref{fig:results_overall_split1}, which summarizes the performance of models trained from scratch, as well as fine-tuned variants of the lower bound, FedFoundAvg, and upper bound methods. Upper Bound consistently shows the best performance across all the downstream tasks. FedFoundAvg closely follows Upper Bound and outperforms both Lower Bound and training from scratch. In Fig. \ref{fig:results_homogeneous} we present the results for different FedFound models for selected metrics.
For classification tasks, the model trained from scratch, and the lower bound achieves the lowest accuracies of 69.8\% and 69.1\% respectively. FedFoundAvg improves classification accuracy over the lower bound by a margin of 8.9\%, reaching 78.0\%, approaching the upper bound accuracy of 81.3\%.
For object detection, the model without pretraining and the lower bound yield mAP values of 10.457 and 14.008.
FedFoundAvg achieves the values of 18.649, an improvement of 4.64 on mAP over the lower bound and narrowing gap with the upper bound (20.860).
For instance segmentation, model without pretraining and the lower bound achieve dice scores of 0.5599 and 0.5202. FedFoundAvg achieves the score of 0.6301, which is again comparable to the upper bound's score of 0.6488. FedFoundAvgM achieves the best dice score (0.6959), followed by FedFoundAdam (0.6617).
We also present the detailed results for all the FedFound models for various metrics for the downstream tasks in Supplementary Table. \ref{tab:performance_split1}.

The results demonstrate that, as expected, the model without pretraining achieves the poorest performance in all the tasks, which is improved moderately by the lower bound, highlighting the importance of pretraining. FedFound models significantly outperform the lower bound. While the upper bound, in general, achieves the best performance across the downstream tasks, FedFound models perform close to the upper bound, demonstrating the potential for FFMs for medical imaging, specifically for gastroendoscopy.  FedFoundAvg consistently demonstrates a stronger performance in the homogeneous setting, outperforming other FL methods in classification and object detection tasks. For segmentation tasks, FedFoundAvgM performs the best. Algorithms like FedFoundAdam and FedFoundAdagrad also perform consistently well on instance segmentation and object detection tasks.

\vspace{-3pt}
\subsection{Performance on Downstream Tasks for Heterogeneous Distribution}
\vspace{-3pt}

\begin{figure}[!t]
	\centerline{\includegraphics[scale=.23]{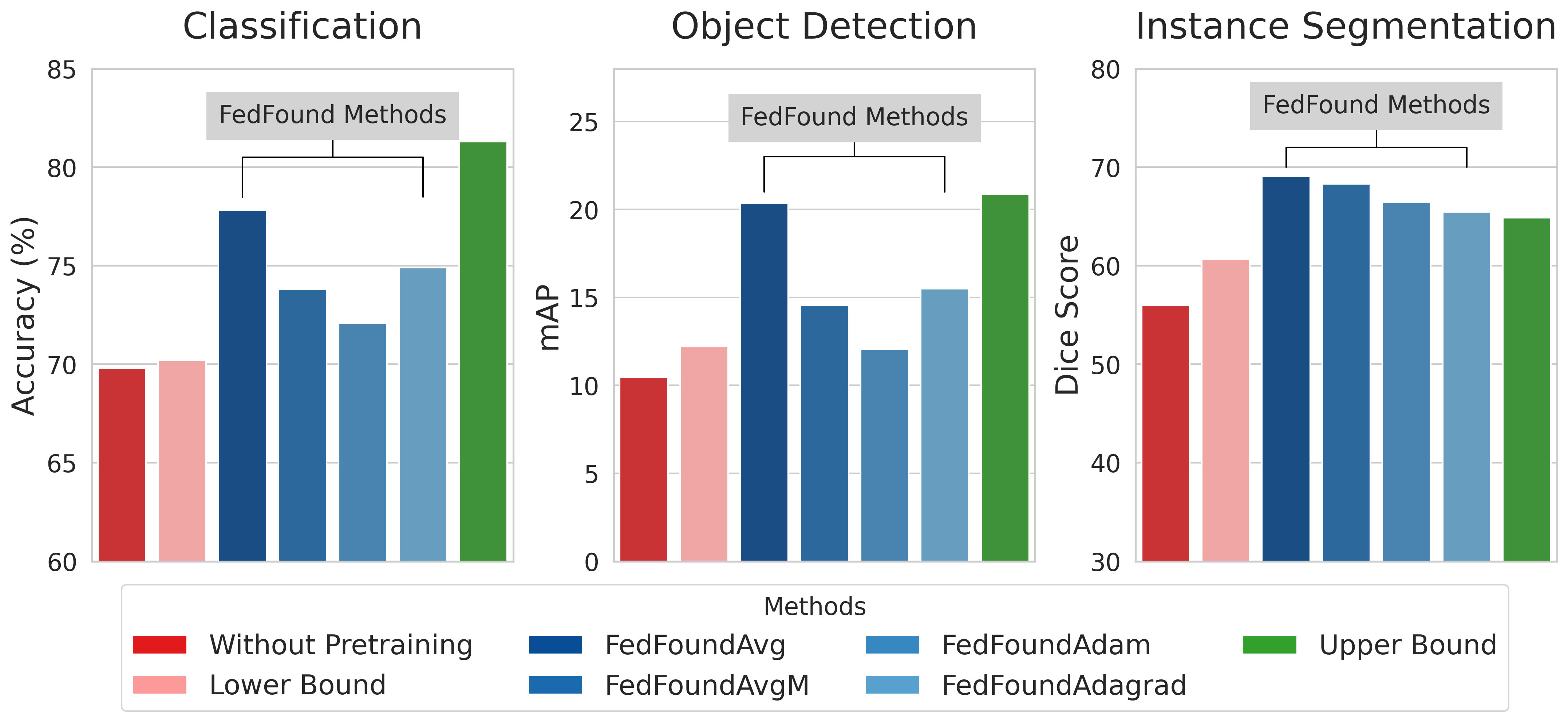}}
	\caption{Performance comparison of various methods on the heterogeneous split. FedFound Models are evaluated against baseline, lower bound, and upper bound models across different downstream tasks.}
	\label{fig:results_heterogeneous}
\end{figure}

Fig. \ref{fig:results_overall_split1} (Right) summarizes the performance of different methods under a heterogeneous setting. Similar to results for homogeneous split, the upper bound performs best across all the tasks, followed by FedFoundAvg. Fig. \ref{fig:results_heterogeneous} presents the results for FedFound models for selected metrics.
It should be noted that the experiments for the model without pretraining and the upper bound are the same for both homogeneous and heterogeneous splits.
For classification, the lower bound achieves the lowest accuracy of 70.2\%. FedFoundAvg improves this by a margin of 7.6\%, reaching 77.8\%, approaching the upper bound accuracy of 81.3\%.
For object detection, the lower bound achieves mAP of 12.227, where FedFoundAvg provides an improvement of 8.126 with mAP value of 20.353. For instance segmentation, the lower bound achieves dice score of 0.6068 and the FedFoundAvg acehives the value of 0.6909, which is even better than the upper bound. FedFoundAvgM achieves the second best dice score of 0.6831. Supplementary Table \ref{tab:performance_split2} provides detailed metrics for all FedFound models under the heterogeneous split.

Similar to homogeneous settings, the FedFound models outperform the lower bound and approach the upper bound, with the FedFoundAvg method demonstrating best performance among them. Its strong performance in heterogeneous distribution for all the downstream tasks marks it as the most effective and robust FL method for developing the FM in real-world applications where data across clients may be non-IID or varied. FedFoundAvgM shows competitive performance for instance segmentation tasks, but falls behind in classification and detection. FedFoundAdagrad and FedFoundAdam provide moderate improvements over the lower bound but don't perform as well as FedFoundAvg or FedFoundAvgM.

In both the homogeneous and heterogeneous splits, models without pretraining and lower bound consistently perform the worst across all tasks. This further emphasizes the importance of pretraining and fine-tuning in the real world and the necessity of FMs. The improvement shown by federated pretraining further highlights the effectiveness of FFMs.

\vspace{-3pt}
\subsection{Impact of Pretraining Epochs on Downstream Task Performance}
\vspace{-3pt}
\begin{figure}[!t]
	\centerline{\includegraphics[scale=.42]{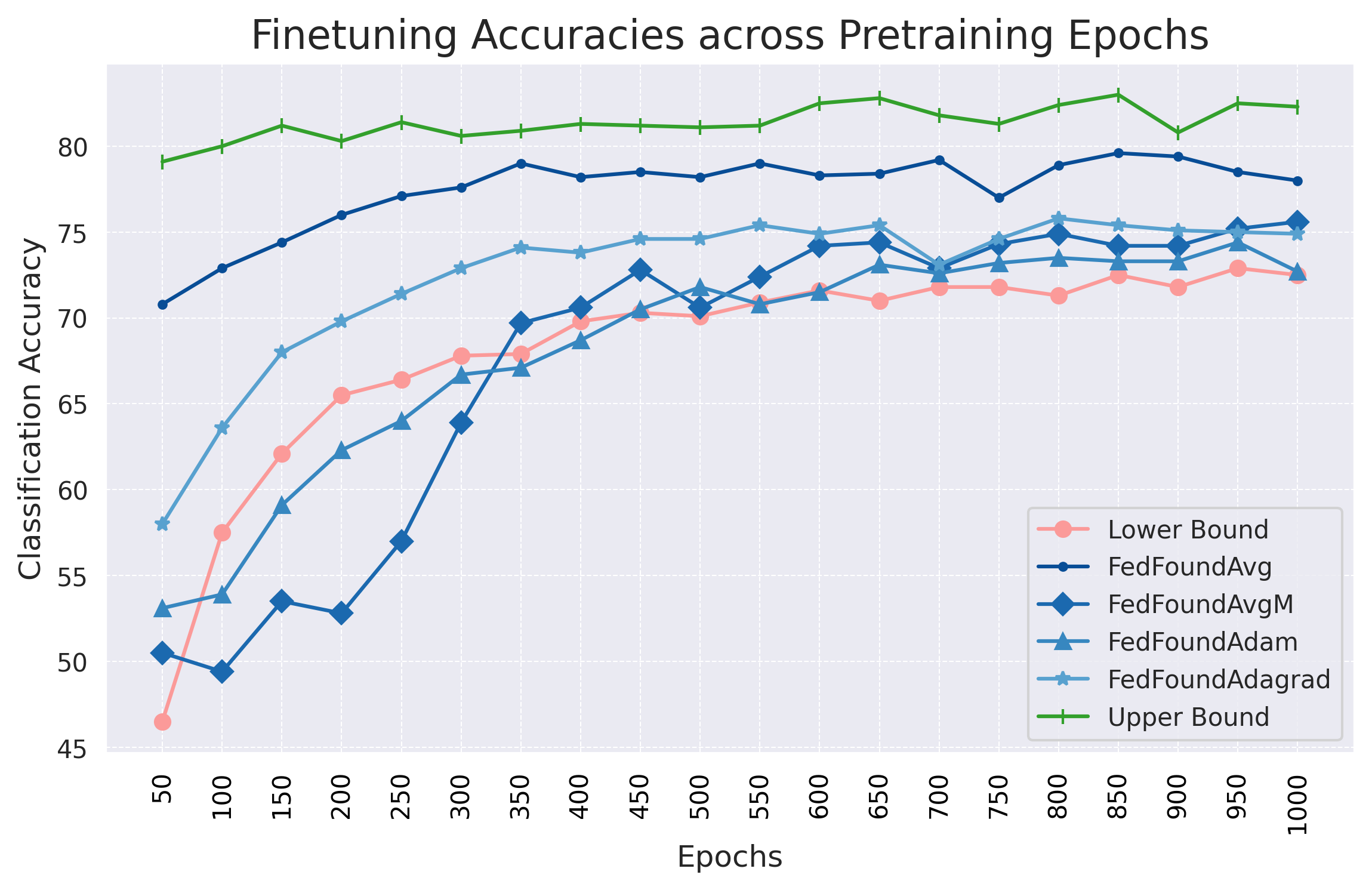}}
	\caption{Finetuning accuracy across pretraining epochs for lower bound, upper bound, and FedFound models. The upper bound shows a plateau early on, while the lower bound increases gradually. FedFoundAvg achieves high initial accuracy and stabilizes quickly.}
	\label{fig:impact_of_epochs}
\end{figure}

Fig. \ref{fig:impact_of_epochs} shows the classification accuracy of the foundation models at different pretraining epochs. The upper bound shows rapid convergence to a stable value, reflecting the capability of centralized training and optimal performance. In contrast, the lower bound rises gradually, indicating slower learning with limited data.
FedFoundAvg shows a steep rise to a stable value with high accuracy from the early epochs, indicating fast and effective learning, though it still lags behind the upper bound. FedFoundAdagrad follows a similar trend with consistently lower accuracy. Other FL algorithms exhibit a pattern similar to the lower bound, surpassing it in later epochs, suggesting slower convergence. These results highlight that FedFoundAvg offers a strong balance, converging faster than other FL methods and approaching centralized performance.

\vspace{-3pt}
\subsection{Classification using Frozen Foundation Model Features}
\vspace{-3pt}
\begin{figure}[!t]
	\centerline{\includegraphics[scale=.35]{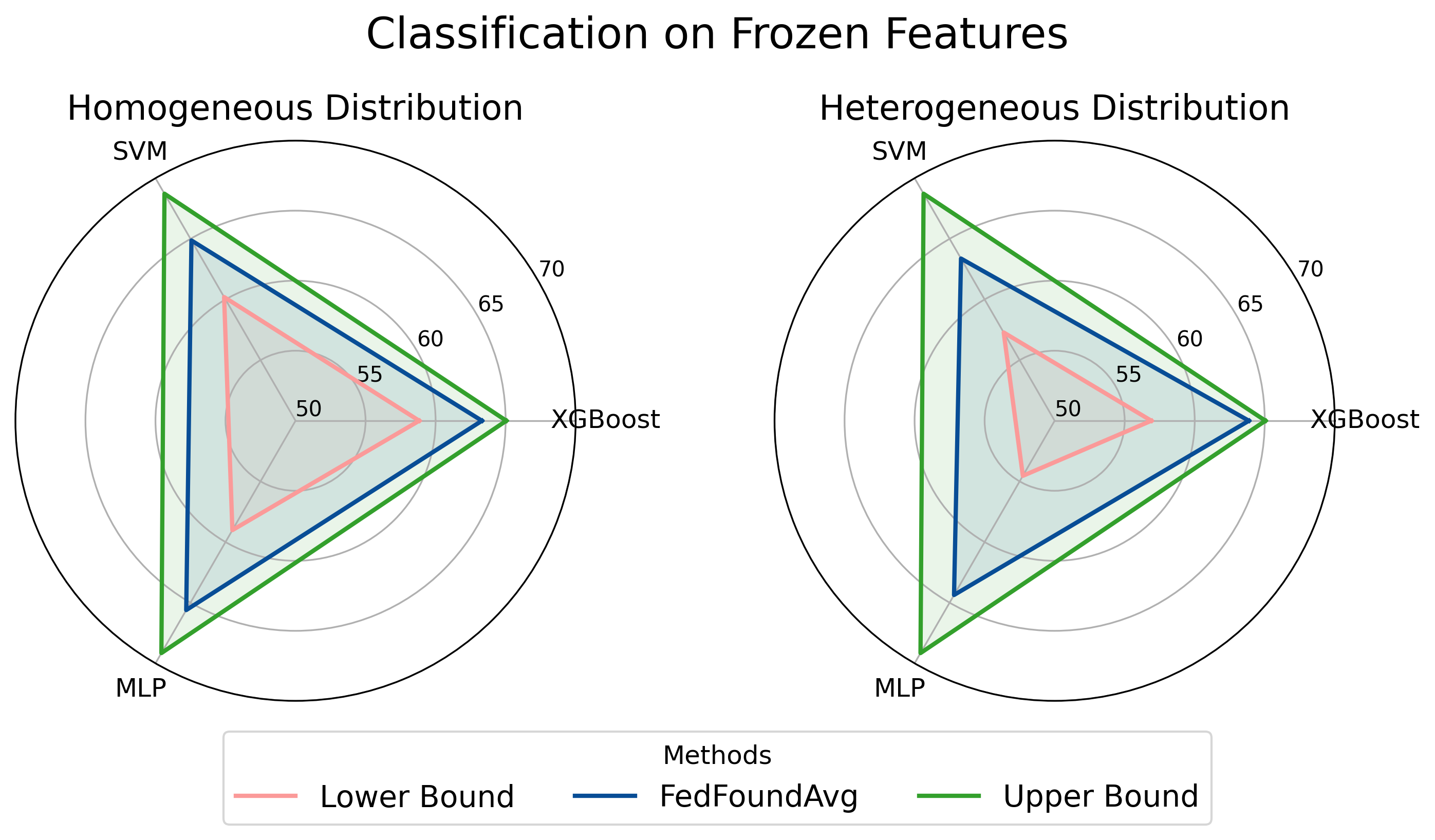}}
	\caption{Evaluation of downstream classification performance using frozen features from various foundation models for homogeneous (Left) and heterogeneous (Right) settings.}
	\label{fig:zero_shot_split1}
\end{figure}

To assess the quality of feature representations learned by the foundation models, we adopt a feature-based transfer learning approach. Features are extracted from the Gastrovision dataset using the lower bound, upper bound, and FedFoundAvg models. We train lightweight classifiers like XGBoost, SVM, and MLP on top of these extracted features. This setup allows for an efficient and lightweight evaluation of the model's generalization capabilities. It provides a practical alternative to end-to-end fine-tuning, particularly when labeled data is limited and training resources are constrained. We train the classifiers on training features, tune hyperparameters using validation features, and evaluate on the test features. 
We apply this strategy in both homogeneous and heterogeneous settings. Fig. ~\ref{fig:zero_shot_split1} shows that FedFoundAvg achieves performance close to the Upper Bound and consistently outperforms the Lower Bound in both settings. Among the evaluated classifiers, MLP achieves the best accuracy with a low training time ($\sim6$ seconds), compared to SVM ($\sim4$ seconds) and XGBoost ($\sim10$ minutes). Detailed results are presented in the Supplementary Figure. ~\ref{fig:zero_shot_mlp}, ~\ref{fig:zero_shot_svm} and ~\ref{fig:zero_shot_xgboost}.

\vspace{-3pt}
\section{Discussion}
\vspace{-3pt}
In this study, we developed FFMs for GI endoscopy imaging, enabling 
collaborative training across multiple sites without sharing sensitive data. 
This approach addresses critical privacy concerns that limit data sharing between hospitals, allowing the development of robust, generalizable models. We trained FFMs using publicly available datasets and evaluated them on three key downstream tasks: classification, object detection, and semantic segmentation.


Our approach integrated state-of-the-art FL optimization techniques, including FedAvg, FedAvgM, FedAdam, and FedAdagrad, into the FFM framework.
We systematically evaluated these methods under both homogeneous and heterogeneous data distributions and demonstrated their performance using standard evaluation metrics.
Our results demonstrate that federated pre-training of FMs significantly improves performance over the models without pretraining or with limited data. In particular, the FedFoundAvg algorithm showed promise in both homogeneous and heterogeneous distributions, closely approaching the performance of the upper bound while maintaining data privacy.

From a clinical perspective, the ability to train high-performing models without centralizing data offers benefits of broader participation from institutions, thus improving generalization and supporting more accurate and consistent detection of GI abnormalities across healthcare centers, potentially resulting in earlier diagnosis and improved patient outcomes, especially in under-resourced healthcare settings.

Despite the observed potential, FFMs still fall short of centralized training performances due to inherent variability across datasets and the inherent challenges of decentralized optimization. Our study was limited to two public datasets, which may not capture the full diversity of the real world. 
Future work shall focus on extending this approach to larger multi-institutional datasets to assess real-world performance.

Overall, this study demonstrates the potential of FL in building robust and high-performance foundation models for medical imaging and can contribute to improved diagnostic accuracy and decision-making in endoscopic procedures.

\section*{Acknowledgment}
Computational resources were provided by the WVU Research Computing Dolly Sods HPC cluster, which is funded in part by NSF OAC-2117575.





\vspace{-3pt}
\section*{References}
\vspace{-3pt}
\bibliographystyle{IEEEtran}
\bibliography{cas-refs}

\clearpage
\FloatBarrier
\onecolumn

\begin{center}
    {\LARGE \bf Supplementary Material}
\end{center}

\appendix

\FloatBarrier

\subsection{FedFound Algorithms}
\renewcommand{\thealgorithm}{A.\arabic{algorithm}}
\setcounter{algorithm}{0}

\begin{algorithm}[H]
\caption{Federated Learning Algorithms}
\begin{algorithmic}[1]
\STATE \textbf{Initialization:} Initial global model $\theta_0$, learning rate $\eta$, momentum coefficient $\alpha$, bias-corrected moments $m_0$, $v_0$, small constant $\epsilon$, set of nodes $S$ containing $N$ clients and a server, number of local data points $n_k$, total data points across all nodes $n$, $\theta_t$ the global model at round $t$, and $\theta_k$ local model at node $k$.
\FOR{$t = 0, \dots, T-1$}
    \STATE $\theta_{t,i,0} = \theta_t$ for each node $i \in S$
    \FOR{each client $i \in S$ in parallel}
        \FOR{$k = 0, \dots, K-1$}
            \STATE Compute an unbiased estimate $g_{t,i,k}$ of $\nabla F_i(\theta_{t,i,k})$
            \STATE $\theta_{t,i,k+1} = \theta_{t,i,k} - \eta g_{t,i,k}$
        \ENDFOR
        \STATE $\Delta_{t,i} = \theta_{t,i,K} - \theta_t$
    \ENDFOR
    \STATE $\Delta_t = \frac{1}{|S|} \sum_{i \in S} \Delta_{t,i}$
    
    \IF{algorithm = `FedAvg'}
        \STATE $\theta_{t+1} = \theta_t - \Delta_t$
    
    \ELSIF{algorithm = `FedAvgM'}
        \STATE $v_{t+1} = \alpha v_t + \Delta_t$
        \STATE $\theta_{t+1} = \theta_t - v_{t+1}$
    
    \ELSIF{algorithm = `FedAdam'}
        \STATE $m_t = \beta_1 m_{t-1} + (1 - \beta_1) \Delta_t$
        \STATE $v_t = \beta_2 v_{t-1} + (1 - \beta_2) \Delta_t^2$
        \STATE $\theta_{t+1} = \theta_t - \eta \frac{\hat{m}_t}{\sqrt{v_t} + \epsilon}$
    
    \ELSIF{algorithm = `FedAdagrad'}
        \STATE $m_t = \beta_1 m_{t-1} + (1 - \beta_1) \Delta_t$
        \STATE $v_t = v_{t-1} + \Delta_t^2$
        \STATE $\theta_{t+1} = \theta_t - \eta \frac{m_t}{\sqrt{v_t + \epsilon}}$
    \ENDIF
\ENDFOR
\end{algorithmic}
\label{algo:fed}
\end{algorithm}

\subsection{Performance on Downstream Tasks}
\FloatBarrier

\renewcommand{\thetable}{B.\arabic{table}}
\setcounter{table}{0}

\subsubsection{Homogeneous Distribution}
\leavevmode

\begin{table}[H]
\centering
\caption{Performance metrics across various downstream tasks with different pretraining methods for homogeneous split. Bold values indicate the highest values for each metric, and underlined values represent the second-highest values.}
\label{tab:performance_split1}
\setlength{\tabcolsep}{5pt}
\renewcommand{\arraystretch}{1.4}
\begin{tabular}{|c|c|c|c|c|c|c|c|}
\hline
\multirow{2}{*}{\textbf{Methods}} & \textbf{Classification} & \multicolumn{3}{c|}{\textbf{Object Detection}} & \multicolumn{3}{c|}{\textbf{Instance Segmentation}} \\
\cline{2-8}
 & \textbf{Accuracy} & \textbf{mAP} & \textbf{mAP50} & \textbf{mAP75} & \textbf{mAP} & \textbf{DSC} & \textbf{JC} \\
\hline
Without Pretraining     & 69.8 & 10.457 & 22.734 & 6.861 & 7.460 & 0.5599 & 0.5548 \\
Lower Bound             & 69.1 & 14.008 & 28.118 & 10.862 & 10.110 & 0.5202 & 0.5153 \\
FedAvg                  & \uline{78.0} & \uline{18.649} & 37.561 & \uline{17.335} & \uline{17.318} & 0.6301 & 0.6239 \\
FedAvgM                 & 76.43 & 16.333 & \uline{43.167} & 9.946 & 5.854 & \textbf{0.6959} & \textbf{0.6894} \\
FedAdam                 & 75.0 & 18.175 & 41.002 & 14.848 & 9.650 & \uline{0.6617} & \uline{0.6564} \\
FedAdagrad              & 75.81 & 15.608 & 34.608 & 10.096 & 8.004 & 0.6510 & 0.6446 \\
Upper Bound             & \textbf{81.3} & \textbf{20.860} & \textbf{45.674} & \textbf{17.442} & \textbf{17.536} & 0.6488 & 0.6435 \\
\hline
\end{tabular}
\end{table}

\clearpage
\subsubsection{Heterogeneous Distribution}
\leavevmode

\begin{table}[H]
\centering
\caption{Performance metrics across various downstream tasks with different pretraining methods for heterogeneous split. Bold values indicate the highest values for each metric, and underlined values represent the second-highest values.}
\label{tab:performance_split2}
\setlength{\tabcolsep}{5pt}
\renewcommand{\arraystretch}{1.4}
\begin{tabular}{|c|c|c|c|c|c|c|c|}
\hline
\multirow{2}{*}{\textbf{Methods}} & \textbf{Classification} & \multicolumn{3}{c|}{\textbf{Object Detection}} & \multicolumn{3}{c|}{\textbf{Instance Segmentation}} \\
\cline{2-8}
 & \textbf{Accuracy} & \textbf{mAP} & \textbf{mAP50} & \textbf{mAP75} & \textbf{mAP} & \textbf{DSC} & \textbf{JC} \\
\hline
Without Pretraining     & 69.8 & 10.457 & 22.734 & 6.861 & 7.460 & 0.5599 & 0.5548 \\
Lower Bound             & 70.2 & 12.227 & 31.213 & 10.447 & 7.499 & 0.6068 & 0.6004 \\
FedAvg                  & \underline{77.8} & \underline{20.353} & \underline{43.307} & \textbf{18.465} & 11.756 & \textbf{0.6909} & \textbf{0.6842} \\
FedAvgM                 & 73.8 & 14.555 & 42.253 & 10.2 & 5.453 & \underline{0.6831} & \underline{0.6775} \\
FedAdam                 & 72.1 & 12.055 & 29.770 & 9.934 & 9.205 & 0.6648 & 0.6592 \\
FedAdagrad              & 74.9 & 15.501 & 34.336 & 12.458 & \underline{11.773} & 0.6546 & 0.6480 \\
Upper Bound             & \textbf{81.3} & \textbf{20.860} & \textbf{45.674} & \underline{17.442} & \textbf{17.536} & 0.6488 & 0.6435 \\
\hline
\end{tabular}
\end{table}

\FloatBarrier
\subsection{Classification using Frozen Foundation Model
Features}
\label{sec:frozen_supple}

\renewcommand{\thefigure}{C.\arabic{figure}}
\setcounter{figure}{0}

\subsubsection{MLP Classifier}
\leavevmode

\begin{figure}[H]
	\centerline{\includegraphics[scale=.23]{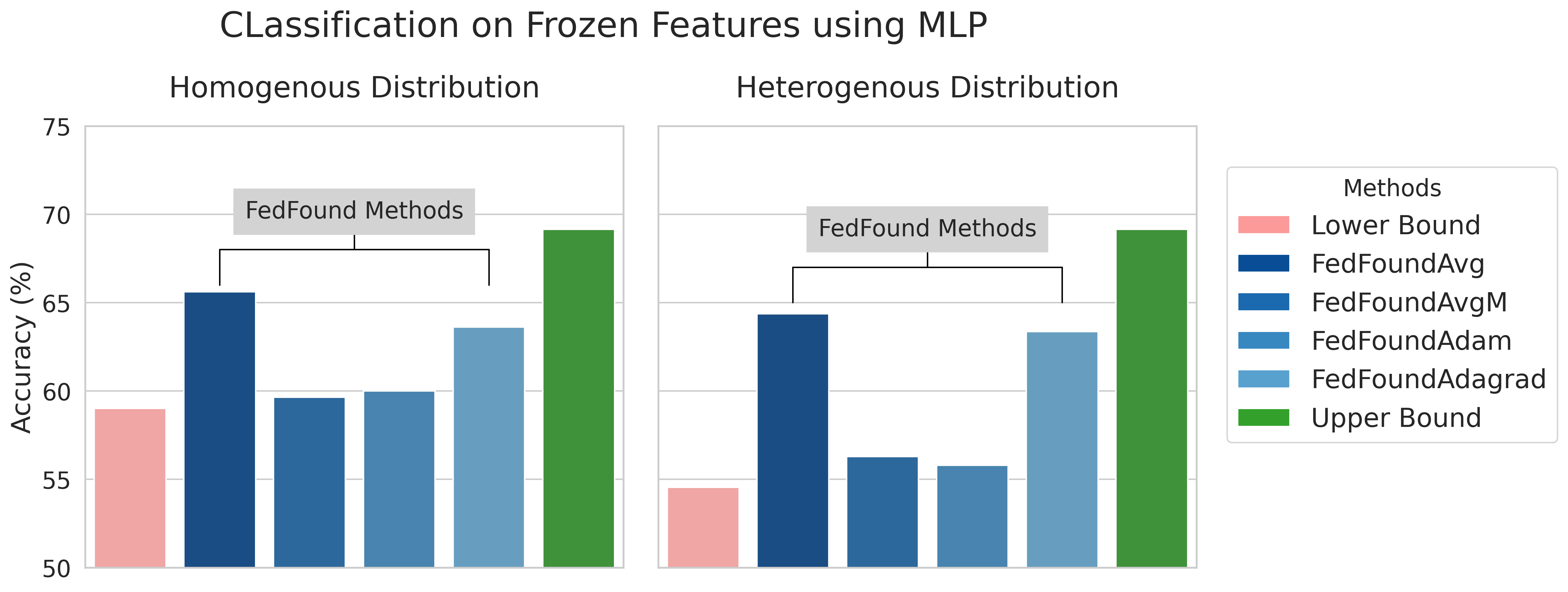}}
	\caption{Downstream classification performance using frozen features from different foundation models under homogeneous (Left) and heterogeneous (Right) settings, using an MLP classifier.}
	\label{fig:zero_shot_mlp}
\end{figure}

\subsubsection{SVM Classifier}
\leavevmode
\begin{figure}[H]
	\centerline{\includegraphics[scale=.23]{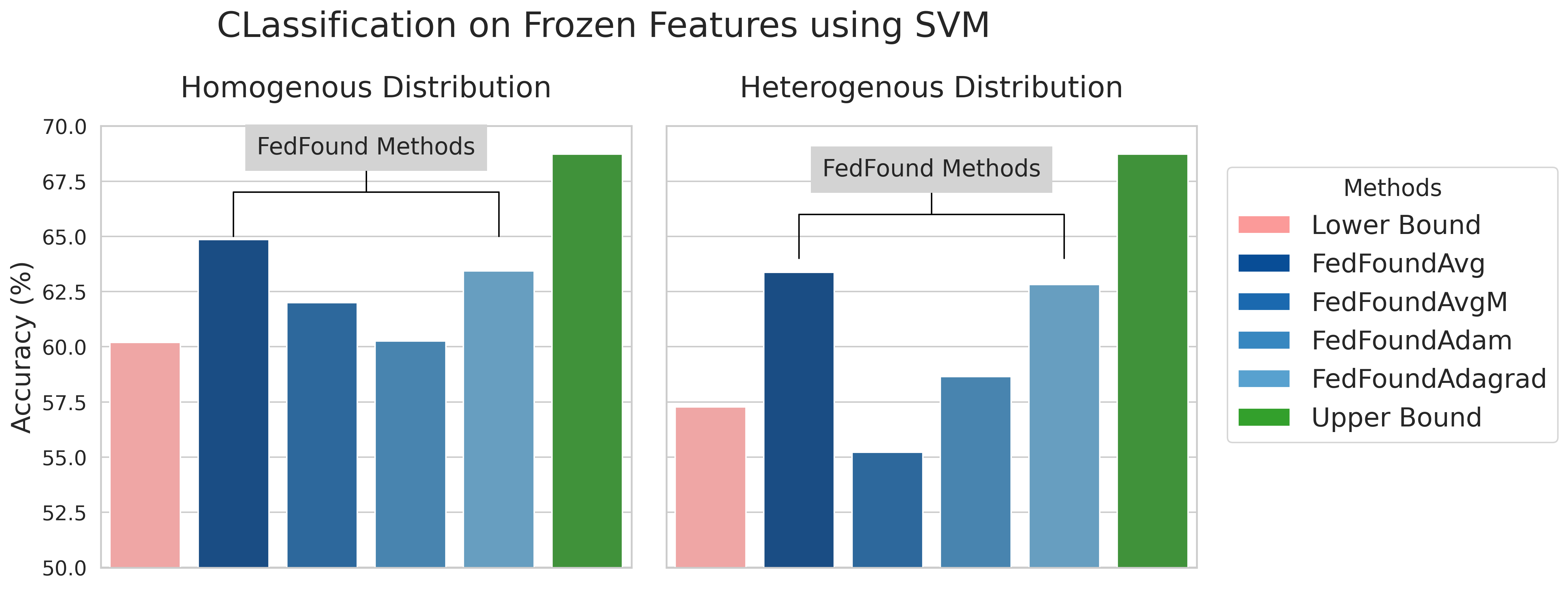}}
	\caption{Downstream classification performance using frozen features from different foundation models under homogeneous (Left) and heterogeneous (Right) settings, using an SVM classifier.}
	\label{fig:zero_shot_svm}
\end{figure}
\FloatBarrier

\subsubsection{XGBoost Classifier}
\leavevmode
\begin{figure}[H]
	\centerline{\includegraphics[scale=.23]{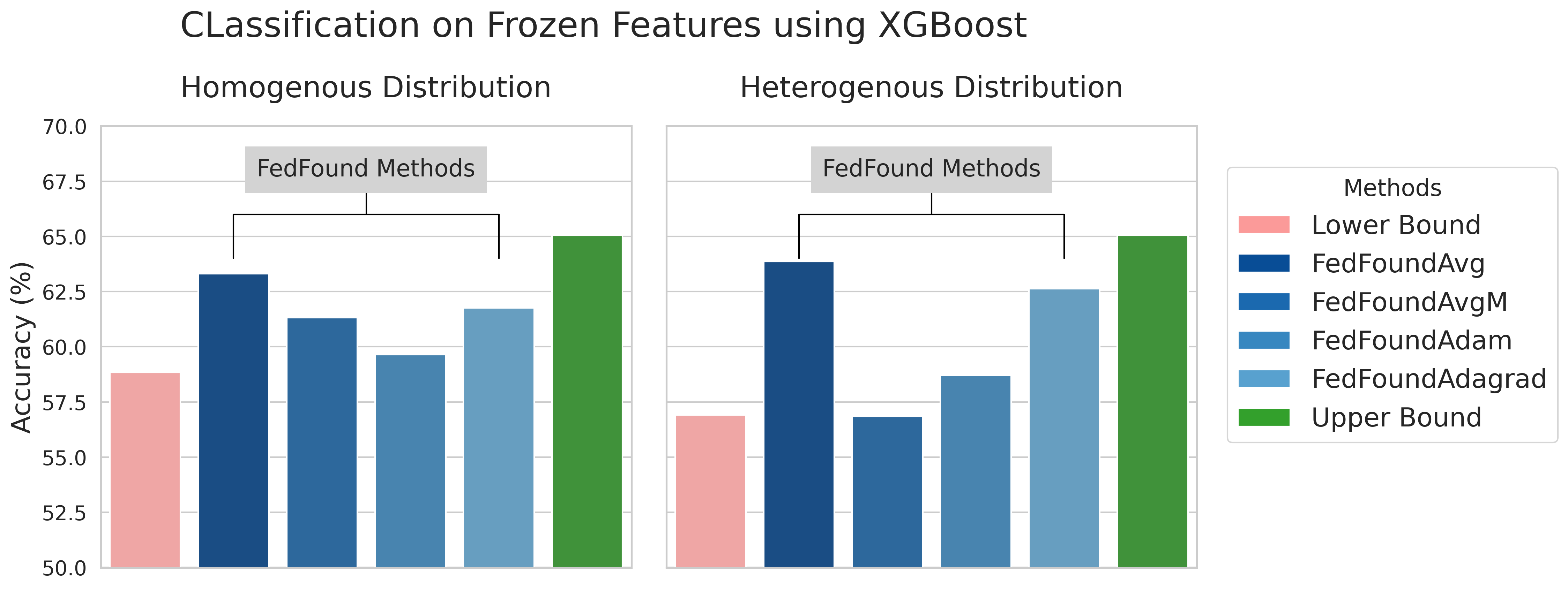}}
	\caption{Downstream classification performance using frozen features from different foundation models under homogeneous (Left) and heterogeneous (Right) settings, using an XGBoost classifier.}
	\label{fig:zero_shot_xgboost}
\end{figure}
\FloatBarrier

\end{document}